\documentclass[10pt]{article} 
\usepackage[preprint]{tmlr}

\usepackage{amsmath,amsfonts,bm}

\def\eqref#1{equation~\ref{#1}}

\def\1{\bm{1}}

\DeclareMathAlphabet{\mathsfit}{\encodingdefault}{\sfdefault}{m}{sl}
\SetMathAlphabet{\mathsfit}{bold}{\encodingdefault}{\sfdefault}{bx}{n}

\DeclareMathOperator*{\argmin}{arg\,min}

\usepackage{hyperref}
\usepackage{url}

\usepackage[small]{caption}
\usepackage{graphicx}
\usepackage{amsmath}
\usepackage{amsthm}
\usepackage{booktabs}

\usepackage[switch]{lineno}

\usepackage{epsfig}
\usepackage{dirtytalk}
\usepackage{placeins}
\usepackage{float}
\usepackage{enumerate}
\usepackage{wrapfig}
\usepackage{enumitem}
\setlist{leftmargin=3mm}
\usepackage{algorithm}
\usepackage{algpseudocode}

\usepackage{multirow, booktabs}

\usepackage{xcolor}

\usepackage{pifont}

\newcommand{\pname}{ADROIT}

\title{ADROIT: A Self-Supervised Framework for Learning \\ Robust Representations for Active Learning}

\author{\name Soumya Banerjee \email soumyab@cse.iitk.ac.in \\
      \addr Department of Computer Science \& Engineering\\
      IIT Kanpur
      \AND
      \name Vinay Kumar Verma \email vinayugc@gmail.com \\
      \addr Amazon, India
      }

\def\month{MM}  
\def\year{YYYY} 
\def\openreview{\url{https://openreview.net/forum?id=XXXX}} 

\begin{document}

\maketitle

\begin{abstract}
  Active learning aims to select optimal samples for labeling, minimizing annotation costs. This paper introduces a unified representation learning framework tailored for active learning with task awareness. It integrates diverse sources, comprising reconstruction, adversarial, self-supervised, knowledge-distillation, and classification losses into a unified VAE-based {\pname} approach. The proposed approach comprises three key components - a unified representation generator (VAE), a state discriminator, and a (proxy) task-learner or classifier. {\pname} learns a latent code using both labeled and unlabeled data, incorporating task-awareness by leveraging labeled data with the proxy classifier. Unlike previous approaches, the proxy classifier additionally employs a self-supervised loss on unlabeled data and utilizes knowledge distillation to align with the target task-learner. The state discriminator distinguishes between labeled and unlabeled data, facilitating the selection of informative unlabeled samples. The dynamic interaction between VAE and the state discriminator creates a competitive environment, with the VAE attempting to deceive the discriminator, while the state discriminator learns to differentiate between labeled and unlabeled inputs. Extensive evaluations on diverse datasets and ablation analysis affirm the effectiveness of the proposed model.
\end{abstract}

\section{Introduction}\label{sec:introduction}

Supervised deep learning~\citep{krizhevsky2012imagenet} has achieved considerable success, particularly in terms of image recognition~\citep{he2016deep}; however, the demand for extensive labeled datasets presents a formidable challenge~\citep{ebrahimi2017gradient}. An alternative approach involves exploring various methods, such as self-supervised representation learning~\citep{gidaris2018unsupervised,chen2020simple}, active learning~\citep{ren2021survey}, and generative models~\citep{kingma2013auto,goodfellow2014generative,luo2022understanding,kingma2021variational}. These diverse methodologies serve distinct purposes; for example, obtaining good performance in self-supervised representation learning requires large amounts of unlabeled data. Generative models like VAEs~\citep{kingma2013auto}, GANs~\citep{goodfellow2014generative}, and diffusion models~\citep{luo2022understanding,kingma2021variational} are typically aimed towards generative tasks rather than representation learning. Active learning focuses on acquiring functions to minimize the number of labeled samples. This study aims to investigate the question: Given access to an expensive labeling oracle, how can we effectively reduce the need for oracle access? While the active learning paradigm addresses this by learning acquisition functions, we propose a more holistic approach. Instead of solely learning an acquisition function, we suggest considering the entire datasets, both labeled and unlabeled with access to a labeling oracle to comprehensively utilize available information, ultimately minimizing the dependency on the labeling oracle.

Numerous active learning approaches, aimed at minimizing labeling costs by selecting informative samples from unlabeled pool~\citep{ren2021survey}, have been proposed. For instance, VAAL~\citep{sinha2019variational} employs both labeled and unlabeled data to model input distributions, selecting informative unlabeled samples. Yet, it is a task-agnostic approach, neglecting the conditional relation between input and output distributions. In contrast, models like TA-VAAL~\citep{kim2021task} and SRAAL~\citep{zhang2020state} incorporate class-conditional relationships, surpassing task-agnostic methods like VAAL~\citep{sinha2019variational} and LL4AL~\citep{yoo2019learning}. Despite these advancements, current methods under-utilize available unlabeled data for task learning~\citep{gao2020consistency,huang2021semi,cabannes2023active}, missing the opportunity to enhance visual representation and model performance~\citep{chen2020simple}. Self-supervised learning~\citep{gidaris2018unsupervised,chen2020simple,chen2021exploring,doersch2015unsupervised,noroozi2016unsupervised} exploits unlabeled data for intermediate representation learning, producing rich and semantically meaningful representations without requiring the knowledge of the annotation space (distribution). However, the integration of self-supervised learning into active learning remains under explored. Our objective is to investigate this integration not only for refining acquisition functions but also for improving the task-learner, unlocking the potential of unlabeled data for both model training and acquisition function learning. 

This paper introduces a novel active learning (AL) method termed \emph{\textbf{\underline{A}} Self-Supervise\textbf{\underline{D}} Framework for Learning \textbf{\underline{RO}}bust Representat\textbf{\underline{I}}ons for Ac\textbf{\underline{T}}ive Learning} ({\pname}). This approach models the input distribution by leveraging both labeled and unlabeled data, integrating self-supervision to enhance representation learning from unlabeled samples. A unified representation learner (VAE) is employed (in the same spirit as VAAL~\citep{sinha2019variational}), with an added adversarial network to identify the most informative unlabeled samples. Task-awareness is achieved through a proxy task-learner, minimizing cross-entropy loss on labeled data to explicitly capture the class-conditional dependence between annotation and input labeled data. The proxy learner also minimizes self-supervised rotation loss (SSL) on unlabeled data, enriching the latent-space representation, enhancing the performance of the proxy as well as the target task-learner, ultimately benefiting both model training and acquisition function learning. Additionally, a teacher-student learning approach is implemented between the target and proxy task-learner, ensuring emulation of the target learner's behavior. Experimental results on diverse benchmark datasets, including balanced and imbalanced ones, demonstrate the superior performance of {\pname} over recent AL baselines. Extensive ablations further confirm the significance of the proposed components.

\noindent\textbf{Our contributions can be summarized as follows:}
\begin{itemize}
    \itemsep0em
    
    \item The proposed model integrates SSL to include both labeled and unlabeled data in the AL model.
    
    \item Utilizing a proxy task-learner or classifier, our model explicitly captures the conditional relationship between the output and input distribution. Additionally, it integrates the self-supervised rotation loss (SSL), making it an SSL-based task-aware AL approach.
    
    \item We employ teacher-student learning between the target and proxy task learner. It enforces the proxy classifier to imitate the behavior of the target model while simultaneously capturing the conditional dependence and minimizing SSL.
    
    \item Empirical evaluations and ablations on diverse datasets with varying class balances validate the superiority of {\pname} over existing baselines.

\end{itemize}

\section{Method}\label{Method}

\subsection{Overview}\label{subsec:overview}

This section outlines the active learning (AL) setup and introduces the notation for the paper. In AL, there's a target task (e.g., image classification) and a task-learner $(T)$ parameterized by $\zeta$. Initially, a large unlabeled pool $X_{U}$ exists, from which $M$ samples are randomly labeled by an oracle, forming the labeled pool $(X_{L}, Y)$. Each AL iteration selects a subset of samples with a budget $(b)$ from $X_{U}$ using an acquisition function and annotates them. Therefore, the labeled pool grows in size with each iteration, whereas the unlabeled pool shrinks. The task-learner $(T)$ is trained in each AL cycle, by minimizing the sum of cross-entropy loss on labeled data and self-supervised rotation-loss~\citep{gidaris2018unsupervised} on unlabeled data. Algorithm~\ref{alg:target_task_learner_training} illustrates the various steps of training the target task-learner $(T_{\zeta})$ in each stage of active learning.

\begin{figure*}[t]
\centering

\includegraphics[width=\textwidth, height=9.5cm]{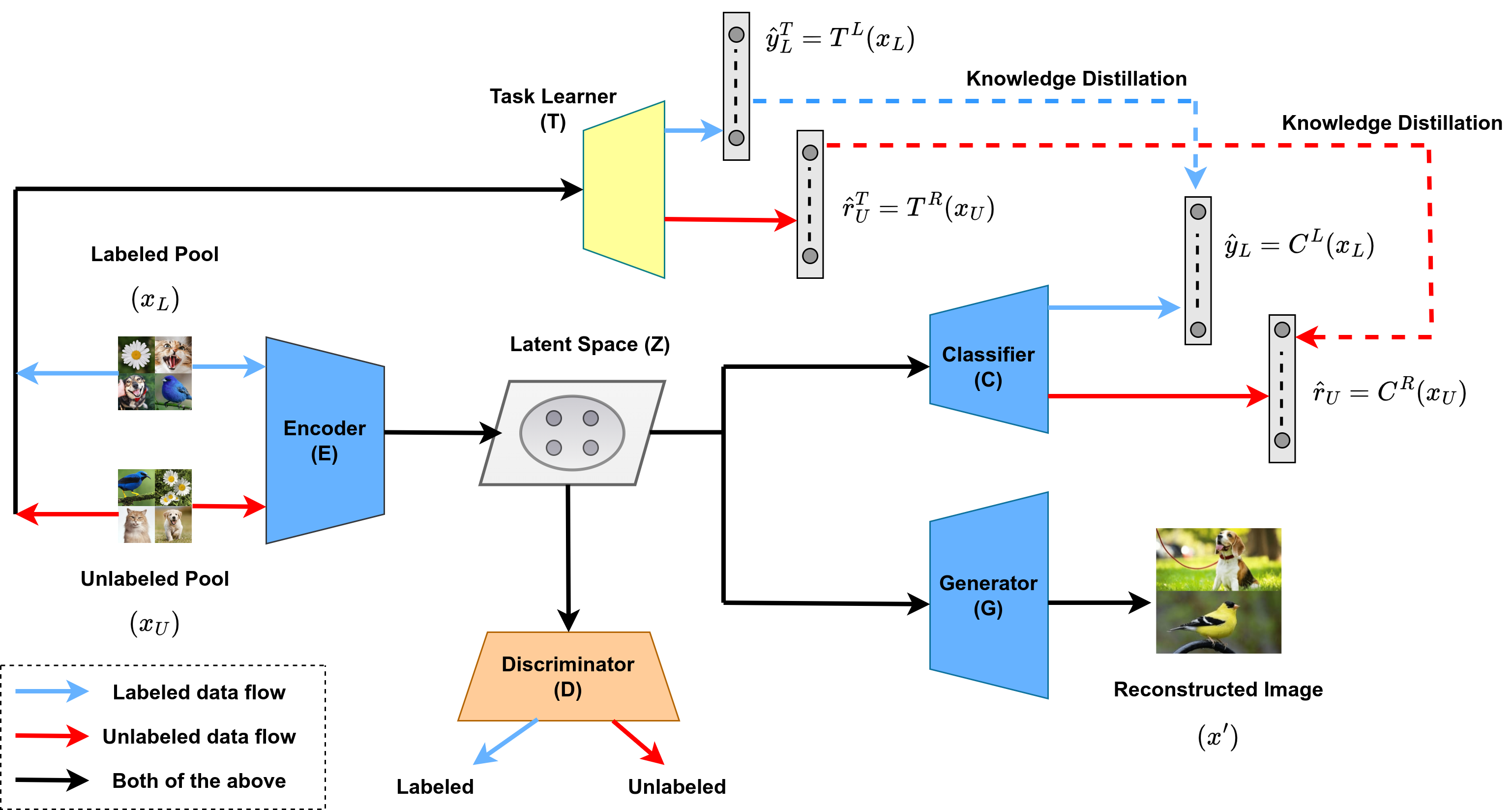}  

\caption{The proposed framework, named \emph{A Self-Supervised Framework for Learning Robust Representations for Active Learning} ({\pname}), comprises a unified representation generator (VAE) with an encoder $(E)$ and a generator network $(G)$, a classifier network $(C)$, and a state discriminator $(D)$. The VAE learns a unified latent space using both labeled and unlabeled data. The classifier network incorporates annotation information into the latent space and refines it through self-supervised loss optimization with unlabeled data. Teacher-student training aligns the classifier $(C)$ with the target task-learner $(T)$. Meanwhile, the state discriminator distinguishes between labeled and unlabeled samples, aiding in the selection of informative unlabeled samples.}
\label{fig:proposed_model}
\end{figure*}

The proposed framework, \emph{A Self-Supervised Framework for Learning Robust Representations for Active Learning} ({\pname}), is illustrated in Figure~\ref{fig:proposed_model}. The subsequent sections provide detailed descriptions of the different components of {\pname}.

\subsection{Unified Representation Learning}\label{subsec:unified_repr_learning}

In this work, we employ a $\beta$-variational autoencoder ($\beta$-VAE)~\citep{kingma2013auto,weng2018VAE} to learn a unified latent-space representation from both labeled and unlabeled data. The encoder network $(E)$, with parameters $\phi$, learns a latent-space representation for the underlying data distribution using a Gaussian prior on the latent code, and the generator network $(G)$, with parameters $\xi$, reconstructs the input data from the latent code. As labeled and unlabeled data share the same distribution $\mathcal{X}$, the VAE learns the latent representation using both of them and is optimized by minimizing the variational lower bound. The objective function for the $\beta$-VAE is formulated as:

\begin{equation}\label{eq:unified_repr_learning}
  \begin{split}
    \mathcal{L}^{\text{URL}}_{\text{VAE}} &= \mathbb{E} \left[ \log G_{\xi} (x_{L}|z_{L}) \right] - \beta\;{\text{D}}_{\text{KL}} \left( E_{\phi}(z_{L}|x_{L}) \left|\right| p(z) \right) 
    + \mathbb{E} \left[ \log G_{\xi} (x_{U}|z_{U}) \right] - \beta\;{\text{D}}_{\text{KL}} \left( E_{\phi}(z_{U}|x_{U}) \left|\right| p(z) \right)
  \end{split}
\end{equation}

where $E_{\phi}(\cdot)$ and $G_{\xi}(\cdot)$ denotes the encoder $(E)$ and the generator $(G)$ network, parameterized by $\phi$ and $\xi$, respectively, $p(z)$ is the prior chosen as the Gaussian unit, and $\beta$ is a hyperparameter to control the trade-off between reconstruction ability and regularization of the latent space.

\subsection{Class Conditional and Self-supervised Representation Learning}\label{subsec:class_cond_self_sup_repr_learning}



In the unified representation learning step (Section~\ref{subsec:unified_repr_learning}), the model captures the input distribution from both labeled and unlabeled data. However, it does not take into account the conditional dependence between inputs and labels. To address this, we introduce a proxy task-learner, mimicking the actual task-learner $(T)$, to incorporate these conditional dependencies into the learning of the latent space.

The proxy task-learner $(C)$, with parameters $\Psi$, serves two purposes. Firstly, it predicts the class labels for the labeled inputs using their latent codes. Secondly, it predicts the rotations applied to unlabeled inputs using their latent codes. It minimizes cross-entropy loss on labeled data and self-supervised rotation-loss (SSL)~\citep{gidaris2018unsupervised} on unlabeled data. This approach explicitly integrates conditional dependence between the inputs and annotations into the latent space while refining the latent representation through SSL. We apply one of six random transformations on unlabeled data and enforces the proxy task-learner to accurately predict the applied transformation: $(i)$ $0^{\circ}$ rotation, $(ii)$ $90^{\circ}$ rotation, $(iii)$ $180^{\circ}$ rotation, $(iv)$ $270^{\circ}$ rotation, $(v)$ horizontal-flip, and $(vi)$ vertical-flip. 


The objective functions for supervised and self-supervised learning are formulated as:

\begin{equation}\label{eq:supervised_learning}
  \mathcal{L}^{L}_{C} = \mathbb{E} \left[ \log C^{L}_{\Psi} (y_{L}|z_{L}) \right] - \text{D}_{\text{KL}} \left( E_{\phi}(z_{L}|x_{L}) \left|\right| p(z) \right)
\end{equation}


\begin{equation}\label{eq:self_sup_learning}
  \mathcal{L}^{U}_{C} = \mathbb{E} \left[ \log C^{R}_{\Psi} (r_{U}|z_{U}) \right] - \text{D}_{\text{KL}} \left( E_{\phi}(z_{U}|x_{U}) \left|\right| p(z) \right)
\end{equation}

where $z_{L}$ and $z_{U}$ denote the latent variable from the latent space for labeled and unlabeled data respectively, $C_{\Psi}(\cdot)$ denotes the proxy task-learner $(C)$, with parameters $\Psi$, and $C^{L}_{\Psi}(\cdot)$ and $C^{R}_{\Psi}(\cdot)$  denotes the labeled and rotation head respectively.

\subsection{Teacher-Student Learning}\label{subsec:teacher_student_learning}

The proxy task-learner $(C)$ aims to mimic the target task-learner $(T)$. Still, it might not accurately replicate how the task-learner estimates the conditional dependence $p(y|x)$ using labeled data. Moreover, the proxy task-learner may learn distinct visual features from those of the task-learner. To align their behavior, we establish teacher-student learning between the target and proxy task-learners, employing knowledge-distillation loss~\citep{hinton2015distilling} where the target task-learner acts as the teacher network. This process involves minimizing the mean-squared error between the corresponding logits generated by the target and proxy task-learner. The target task-learner remains unchanged during this optimization, with only the proxy task-learner being updated.

The objective function for teacher-student learning is formulated as:

\begin{equation}\label{eq:teacher_student_learning}
  \begin{split}
    \mathcal{L}_{\text{KD}} &= \mathbb{E}_{x_{L} \sim X_{L}} \left[ \left|\right| T_{\zeta}^{L}(x_{L}) - C^{L}_{\Psi}(E_{\phi}(x_{L})) \left|\right|_{2}^{2} \right] 
    + \mathbb{E}_{x_{U} \sim X_{U}} \left[ \left|\right| T^{R}_{\zeta}(x_{U}) - C^{R}_{\Psi}(E_{\phi}(x_{U})) \left|\right|_{2}^{2} \right]
  \end{split}
\end{equation}

where $X_{L}$ and $X_{U}$ denote labeled and unlabeled data pool respectively, $E_{\phi}(\cdot)$ denote the encoder network, $C_{\Psi}(\cdot)$ and $T_{\psi}(\cdot)$ represents proxy and target task-learner respectively, and $T^{L}_{\zeta}(\cdot)$ and $T^{R}_{\zeta}(\cdot)$ denote the labeled and rotation head, respectively.

\begin{algorithm}[t] 
  \caption{{\pname}}\label{alg:proposed_model}
  \begin{algorithmic}[1]

  \Require Labeled pool $(X_{L}, Y)$, Unlabeled pool $(X_{U})$

  \Require Trained task-learner $T_{\zeta}(\cdot)$ \Comment{it is kept frozen while training the VAE and the state-discriminator}

  \Require Initialize model parameters: $\Theta_{\text{VAE}}$, and $\theta_{D}$ \Comment{$\Theta_{\text{VAE}}$ represents the parameters of the whole VAE network including Encoder $(E)$, Generator $(G)$, \& Classifier $(C)$, and $\theta_{D}$ represents the parameters of the state-discriminator $(D)$}
  \Require Hyper-parameters: epochs, $\lambda_{1}, \lambda_{2}, \lambda_{3}, \lambda_{4}, \alpha_{1}$, and $\alpha_{2}$
  
  \For{$e = 1$ to epochs} 

    \State sample $(x_{L}, y) \sim (X_{L}, Y)$ and $x_{U} \sim X_{U}$


    \State Compute $\mathcal{L}_{\text{VAE}}^{\text{URL}}$ by using Eq.~(\ref{eq:unified_repr_learning})

    \State Compute $\mathcal{L}_{C}^{L}$ by using Eq.~(\ref{eq:supervised_learning})

    \State Compute $\mathcal{L}_{C}^{U}$ by using Eq.~(\ref{eq:self_sup_learning})

    \State Compute $\mathcal{L}_{\text{KD}}$ by using Eq.~(\ref{eq:teacher_student_learning})

    \State Compute $\mathcal{L}_{\text{VAE}}^{\text{adv}}$ by using Eq.~(\ref{eq:vae_as_generator})

    \State $\mathcal{L}_{\text{VAE}} \leftarrow \mathcal{L}^{\text{URL}}_{\text{VAE}} + \lambda_{1} \; \mathcal{L}^{L}_{C} + \lambda_{2} \; \mathcal{L}^{U}_{C} + \lambda_{3} \; \mathcal{L}_{\text{KD}} + \lambda_{4} \; \mathcal{L}^{\text{adv}}_{\text{VAE}}$

    \State Update VAE by descending its stochastic gradients:

    \State $\Theta_{\text{VAE}}^{\prime} \leftarrow \Theta_{\text{VAE}} - \alpha_{1} \nabla \mathcal{L}_{\text{VAE}}$

    \State Compute $\mathcal{L}_{D}$ by using Eq.~(\ref{eq:discriminator})

    \State Update $D$ by descending its stochastic gradients:

    \State $\theta_{D}^{\prime} \leftarrow \theta_{D} - \alpha_{2} \nabla \mathcal{L}_{D}$

  \EndFor

  \State \Return Trained $\Theta_{\text{VAE}}$, and $\theta_{D}$

  \end{algorithmic}
\end{algorithm}

\subsection{Adversarial Representation Learning}\label{subsec:adversarial_repr_learning}

The VAE's latent space integrates both labeled and unlabeled data, capturing input distribution $p(x)$ and conditional dependence $p(y|x)$. Our objective is to leverage these representations for identifying informative samples from the unlabeled pool. We achieve this by training an adversarial network on the VAE's latent space, akin to a GANs' discriminator~\citep{goodfellow2014generative}. It discerns labeled and unlabeled latent codes (treating it as a classification problem where the labeled inputs are class 1 and unlabeled inputs are class 0), effectively categorizing the latent space into these two groups. Once trained, the adversarial network assists in selecting informative unlabeled samples: unlabeled inputs that ``look like'' labeled inputs, i.e., high probability of belonging to class 1, are discarded, and (a subset of the) others are chosen for annotation by the oracle (using a scheme in Section~\ref{subsec:sampling_strategy}).

The VAE and adversarial network are jointly trained adversarially, similar to GANs~\citep{goodfellow2014generative}. The VAE emulates the generator, aiming to mislead the adversarial network into classifying both labeled and unlabeled inputs as 1. Conversely, the adversarial network correctly classify latent representations as labeled (class 1) or unlabeled (class 0), distinguishing between samples from the labeled pool $X_{L}$ and the unlabeled pool $X_{U}$.

The objective function for VAE as the generator network is formulated as the binary cross-entropy loss as:

\begin{equation}\label{eq:vae_as_generator}
  \mathcal{L}^{\text{adv}}_{\text{VAE}} = - \mathbb{E} \left[ \log D_{\theta} (E_{\phi}(z_{L}|x_{L})) \right] - \mathbb{E} \left[ \log D_{\theta} (E_{\phi}(z_{U}|x_{U})) \right]
\end{equation}

where $D_{\theta}(\cdot)$ represents the adversarial discriminator network $(D)$, parameterized by $\theta$.

The objective function for adversarial learning of the discriminator is formulated as:

\begin{equation}\label{eq:discriminator}
  \mathcal{L}_{D} = - \mathbb{E} \left[ \log D_{\theta} (E_{\phi}(z_{L}|x_{L})) \right] - \mathbb{E} \left[ \log (1 - D_{\theta} (E_{\phi}(z_{U}|x_{U}))) \right]
\end{equation}

The total objective function for the VAE is obtained by combining Eq.~(\ref{eq:unified_repr_learning}), Eq.~(\ref{eq:supervised_learning}), Eq.~(\ref{eq:self_sup_learning}), Eq.~(\ref{eq:teacher_student_learning}) and Eq.~(\ref{eq:vae_as_generator}), as follows:

\begin{equation}\label{eq:vae_total_objective}
  \mathcal{L}_{\text{VAE}} = \mathcal{L}^{\text{URL}}_{\text{VAE}} + \lambda_{1} \; \mathcal{L}^{L}_{C} + \lambda_{2} \; \mathcal{L}^{U}_{C} + \lambda_{3} \; \mathcal{L}_{\text{KD}} + \lambda_{4} \; \mathcal{L}^{\text{adv}}_{\text{VAE}}
\end{equation}

where $\lambda_{1}, \lambda_{2}, \lambda_{3}$ and $\lambda_{4}$ are hyperparameters that determine the effect of various components to learn an effective latent-space representation.

Algorithm~\ref{alg:proposed_model} shows various steps of the proposed model.

\begin{algorithm}[t] 
  \caption{Sampling Strategy}\label{alg:sampling_strategy}
  \begin{algorithmic}[1]
  
  \Require $(X_{L}, Y)$, and $X_{U}$

  \Require Trained models: $\Theta_{\text{VAE}}$ and $\theta_{D}$ \Comment{$\Theta_{\text{VAE}}$ denotes the parameters of the whole VAE and $\theta_{D}$ denotes the parameters of the state-discriminator $(D)$}

  \Require Sampling budget: $b$
  
  \State Select samples $(X_{s})$ with $\text{min}_{b}\{ D_{\theta}(E_{\phi} (z_{U}|x_{U})) \}$

  \State $Y_{o} \leftarrow \mathcal{ORACLE}(X_{s})$

  \State $(X_{L}, Y) \leftarrow (X_{L}, Y) \cup (X_{s}, Y_{o})$

  \State $X_{U} \leftarrow X_{U} - X_{s}$ 

  \State \Return $(X_{L}, Y)$ and $X_{U}$

  \end{algorithmic}
\end{algorithm}

\begin{algorithm}[t] 
  \caption{Target Task-Learner Training}\label{alg:target_task_learner_training}
  \begin{algorithmic}[1]
  
  \Require $(X_{L}, Y)$, and $X_{U}$

  \Require Target Task-learner: $T_{\zeta}{(\cdot)}$ 

  \Require Initialize model parameters: $\zeta_{T}$ \Comment{$\zeta_{T}$ denotes the parameters of the target task-learner $(T)$}

  \Require Hyper-parameters: epochs, $\eta$, $\xi$
  
  \For{$e = 1$ to epochs}

    \State sample $(x_{L}, y) \sim (X_{L}, Y)$ and $x_{U} \sim X_{U}$

    \State $\mathcal{L}^{L}_{T} = \mathcal{L}_{CE}(y, T^{L}_{\zeta}(x_{L}))$ \Comment{$\mathcal{L}_{CE}(\cdot, \cdot)$ denotes cross-entropy loss and $T^{L}_{\zeta}(\cdot)$ represents that the label-prediction head is used to generate output}

    \State $\mathcal{L}^{U}_{T} = \mathcal{L}_{CE}(r_{U}, T^{R}_{\zeta}(x_{U}))$ \Comment{$T^{R}_{\zeta}(\cdot)$ represents that the rotation-prediction head is used to generate output}

    \State $\mathcal{L} = \mathcal{L}^{L}_{T} + \xi \; \mathcal{L}^{U}_{T}$

    \State Evaluate $\nabla_{\zeta}\mathcal{L}(T_{\zeta})$

    \State Update task-learner $(T)$ by descending its stochastic gradients:

    \State $\zeta_{T}^{\prime} = \zeta_{T} - \eta \; \nabla_{\zeta}\mathcal{L}(T_{\zeta})$

  \EndFor

  \State \Return Trained Task-learner $T_{\zeta}(\cdot)$

  \end{algorithmic}
\end{algorithm}

\subsection{Sampling Strategy}\label{subsec:sampling_strategy}

After training {\pname}, the data-points $(x^{1}_{U}, \dots, x^{b}_{U})$ to be labeled at each iteration are selected as:

\begin{equation}\label{eq:sampling_strategy}
  (x^{1}_{U}, \dots, x^{b}_{U}) = \argmin_{(x^{1}_{U}, \dots, x^{b}_{U}) \subset X_{U}} D_{\theta}(E_{\phi}(z_{U}|x_{U})) 
\end{equation}

The details of sample selection is shown in the Algorithm~\ref{alg:sampling_strategy}.

\section{Related Work}\label{sec:related_work}

\subsection{Active Learning}\label{subsec:active_learning}

Active Learning (AL)~\citep{ren2021survey,sinha2019variational,kim2021task,mottaghi2019adversarial,zhang2020state,guo2021dual,wang2020dual,shui2020deep,li2021learning,jin2022cold,ebrahimi2020minimax,sener2017active,agarwal2020contextual,geifman2019deep,gal2017deep,kirsch2019batchbald,yoo2019learning,zhan2022comparative} strategies aim to choose the most informative samples from an unlabeled pool for labeling by an oracle. Recent works have proposed various AL methods, which broadly can be categorized into $(i)$ task-agnostic and $(ii)$ task-aware approaches, depending on whether the AL sample selection strategy explicitly considers the conditional relationship between the output and input distribution. Moreover, existing AL techniques can also be grouped into $(i)$ uncertainty-based and $(ii)$ diversity-based approaches. Some AL methods are also characterized as query-by-committee-based approaches. The following provides a brief overview of these diverse AL approaches.

\subsubsection{Task-agnostic and Task-aware AL approaches}\label{subsubsec:task_agnostic_aware_al}

Task-agnostic approaches~\citep{sinha2019variational,ren2021survey,zhu2017generative,yoo2019learning} model the input data distribution without considering the conditional relationship between input and output. VAAL~\citep{sinha2019variational} models the input distribution by employing data from both the labeled and unlabeled pools through a VAE~\citep{kingma2013auto}. The VAE captures a low-dimensional latent space representation, which is subsequently used to train an adversarial network~\citep{goodfellow2014generative}. Once trained, this adversarial network identifies the most informative samples. Another example is GAAL~\citep{zhu2017generative}, which generates adversarial samples causing high uncertainty in the learner. These samples are then labeled by an oracle and integrated into the labeled pool. Similarly, LL4AL~\citep{yoo2019learning} learns to predict target losses for unlabeled inputs and selects unlabeled samples with the top-k losses for labeling and inclusion in the labeled pool. In contrast, task-aware methods~\citep{ren2021survey,kim2021task,zhang2020state} explicitly leverage conditional dependence through task-specific modeling. SRAAL~\citep{zhang2020state}, an extension of VAAL~\citep{sinha2019variational}, introduces an uncertainty indicator for unlabeled samples, assigning them different levels of importance. TA-VAAL~\citep{kim2021task} incorporates a ranking-loss framework~\citep{yoo2019learning,saquil2018ranking} into the VAE structure. Both SRAAL~\citep{zhang2020state} and TA-VAAL~\citep{kim2021task} employ an adversarial network to identify the top-k unlabeled samples.

\subsubsection{Uncertainty and Diversity based AL approaches}\label{subsubsec:uncertainty_diversity_al}


The distinction between uncertainty-based and diversity-based AL approaches lies in their selection of unlabeled samples. Uncertainty-based strategies, like BALD~\citep{gal2017deep} and MC-Dropout~\citep{houlsby2011bayesian}, target samples where the model is uncertain, using this uncertainty as an indicator of potential model errors. In contrast, diversity-based approaches, such as Coreset~\citep{sener2017active}, aim to maximize diversity within the selected samples. For example, BALD maximizes mutual information between model predictions and parameters, while Coreset employs a greedy $K$-center algorithm for diverse sample selection. Additionally, DAAL~\citep{wang2020dual} utilizes two adversarial networks, with one estimating uncertainty and the other maximizing the diversity of chosen samples.

\subsubsection{Query-By-Committee based AL approaches}\label{subsubsec:querry_by_committee_al}


\say{Query-By-Committee} based AL approaches, also referred to as agreement-based AL methods, diverge from the conventional AL approaches by utilizing an ensemble of diverse models to gauge the uncertainty of a sample from the unlabeled pool rather than relying on a single model~\citep{beluch2018power,cohn1994improving,cortes2019active,iglesias2011combining,mccallum1998employing,seung1992query,gao2020consistency}. For a sample, if the predictions across the ensemble of models vary largely then this sample is considered most uncertain sample. Therefore, it is selected for labeling.

\subsection{Self-supervised Learning}\label{subsec:self_supervised_learning}

Self-supervised learning (SSL)\citep{gidaris2018unsupervised,chen2020simple,mishra2022simple,he2022masked,chen2021exploring,caron2021emerging,li2021efficient,weng2019selfsup,balestriero2023cookbook} aims to train deep neural networks (DNNs) using almost freely available vast amounts of unlabeled data by defining a pretext task based on the data itself. This allows the model to learn rich, descriptive, and generic representations for downstream tasks. In this paper, we applied six different random transformations uniformly to the unlabeled samples and enforced the network to predict the applied transformation as a pretext task. This approach aims to learn robust visual features using the freely available, abundant pool of unlabeled data (refer to Section~\ref{subsec:class_cond_self_sup_repr_learning}).

\subsection{Variational AutoEncoder}\label{subsec:variational_autoencoder}

Autoencoders~\citep{hinton2006reducing,hinton2011transforming} have long been used for representation learning and dimensionality reduction. They comprise two networks: an encoder, which maps input data to a latent space, and a decoder, which generates input data from the latent code. These networks learn identity mapping in an unsupervised manner by minimizing reconstruction errors. The Variational Autoencoder (VAE)~\citep{kingma2013auto,weng2018VAE,doersch2016tutorial} is also a latent variable model that follows a similar encoder-decoder architecture. However, a VAE's encoder network maps each high-dimensional input data to a latent distribution using a Gaussian prior for the latent code. The decoder then reconstructs the input data from this latent distribution. Once trained, the decoder can operate independently to generate new synthetic data. Recently, a lot of research has exploited VAE-like architectures in various applications, ranging across generative modeling~\citep{kingma2013auto,chen2016variational,kingma2021variational,kingma2016improved,vahdat2020nvae,razavi2019generating,yan2021videogpt}, anomaly detection~\citep{xu2018unsupervised}, semi-supervised learning~\citep{kingma2014semi}, sequence-to-sequence modeling~\citep{kaiser2018fast,bahuleyan2017variational}, and active learning~\citep{sinha2019variational,ren2021survey}. In this paper, we use a VAE-like architecture to learn a unified latent representation by exploiting both labeled and unlabeled data at each stage of active learning (refer to Section~\ref{subsec:unified_repr_learning}).

\subsection{Adversarial Learning}\label{subsec:adversarial_learning}

The Generative Adversarial Network (GAN)\citep{goodfellow2014generative} comprises two networks: (i) a generator network that learns to create realistic data from a latent Gaussian distribution, and (ii) a discriminator network that learns to accurately differentiate between the generated (fake) samples and real data. Adversarial learning has been extensively used in various domains, including generative modeling\citep{goodfellow2014generative,radford2015unsupervised,mirza2014conditional,wang2022diffusion}, image-to-image translation~\citep{zhu2017unpaired,lin2018conditional,isola2017image,ma2018gan}, domain generalization~\citep{ganin2015unsupervised,zhang2022generalized,huang2018auggan}, data augmentation~\citep{antoniou2017data,zhu2017data,calimeri2017biomedical}, representation learning~\citep{chen2016infogan,donahue2017semantically,huang2017stacked,odena2016semi,donahue2016adversarial}, and active learning~\citep{sinha2019variational,ren2021survey}. In this paper, we also exploit adversarial learning to determine the most informative unlabeled samples for annotation at each stage of active learning (refer to Section~\ref{subsec:adversarial_repr_learning}).

\section{Experiments}\label{sec:experiments}

We conduct extensive experiments to demonstrate the effectiveness of our proposed approach. For a robust evaluation, we evaluate our approach on class-balanced as well as class-imbalanced datasets.

\begin{figure}[t]

  \centering
  \includegraphics[width=\textwidth, height=14cm]{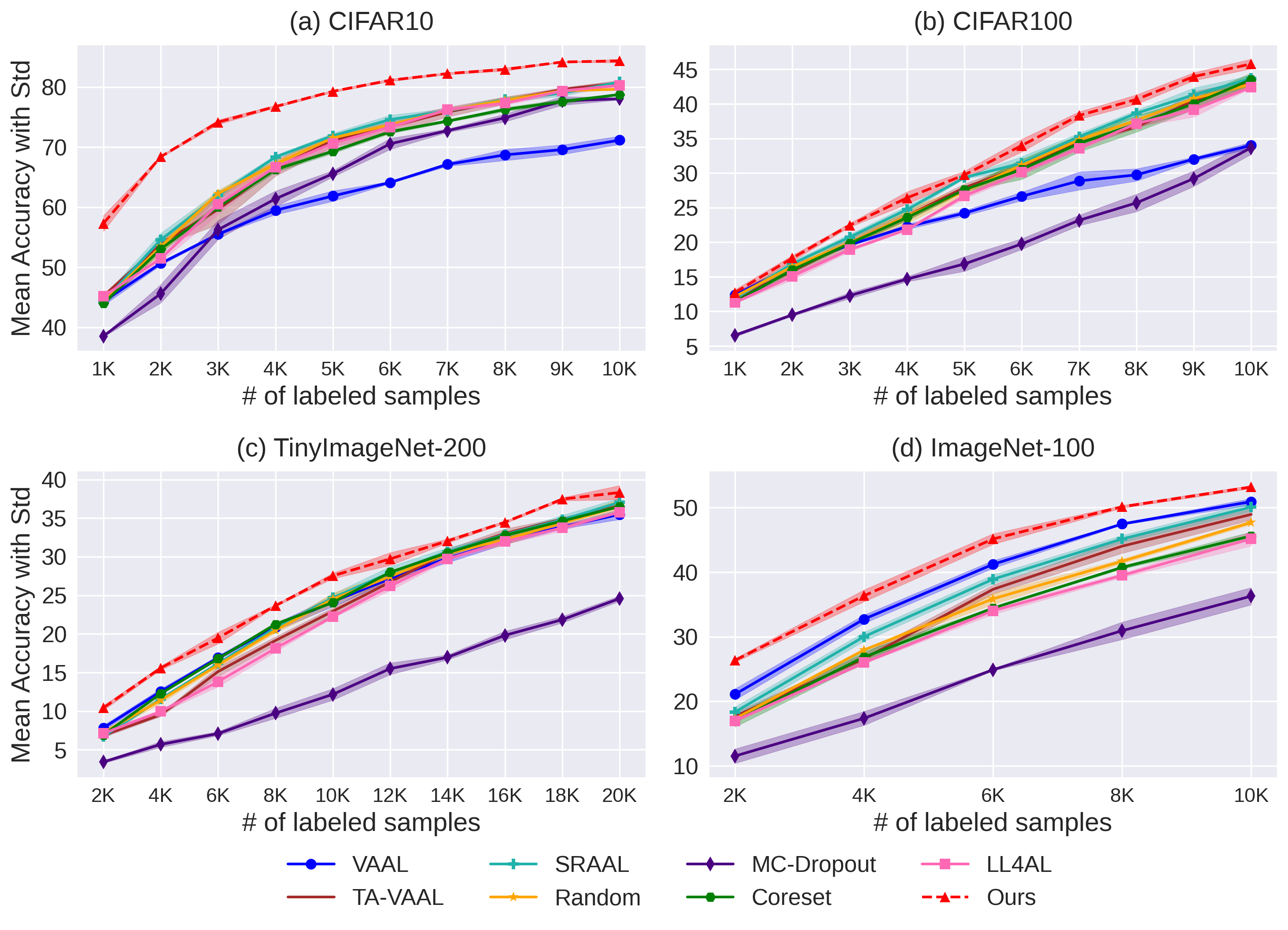}  

  \caption{Mean accuracy with standard deviations (shaded) of AL methods over the number of labeled samples on CIFAR10/100 (balanced), TinyImageNet-200 (balanced) and ImageNet-100 (balanced).}
  \label{fig:balanced_datasets}
\end{figure}

\subsection{Active Learning on Balanced Datasets}\label{subsec:balanced_dataset}

\textbf{Balanced benchmark datasets.} We evaluate the proposed AL model ({\pname}) on four balanced benchmark datasets: CIFAR10/100~\citep{krizhevsky2009learning}, TinyImageNet-200~\citep{le2015tiny} and ImageNet-100 (a subset of ImageNet-1K~\citep{deng2009imagenet}). Initially, 1000 and 2000 samples are randomly labeled for CIFAR10/100 and TinyImageNet-200/ImageNet-100. In subsequent iterations, 1000 and 2000 samples are selected through active learning strategies and annotated by an oracle. {We provide more details in the appendix.}

\textbf{Compared methods.} We compare the performance of our method ({\pname}) with recent state-of-the-art baselines, including VAAL~\citep{sinha2019variational}, MC-Dropout~\citep{gal2017deep}, Coreset~\citep{sener2017active}, LL4AL~\citep{yoo2019learning}, SRAAL~\citep{zhang2020state}, and TA-VAAL~\citep{kim2021task}. Random sample selection is also included for comparison. All methods are trained from scratch using original code and hyperparameters, and evaluation employs the same target learner.

\textbf{Implementation details.} We use ResNet-18~\citep{he2016deep} as the target task-learner for ImageNet-100, along with adapting it for CIFAR10/100 and TinyImageNet-200, with SGD optimizer (lr: $0.01$, momentum: $0.9$, weight decay: $0.005$). The VAE employs a modified Wasserstein autoencoder~\citep{tolstikhin2017wasserstein} with a 5-layer MLP discriminator, similar to VAAL~\citep{sinha2019variational} and TA-VAAL~\citep{kim2021task}. Both the VAE and discriminator use the Adam optimizer~\citep{kingma2014adam} (lr: $5\times10^{-4}$). {For more details, please refer to the appendix.}





\textbf{Performance on CIFAR10.} In Figure~\ref{fig:balanced_datasets}(a), we present the performance of active learning (AL) methods, including {\pname}, on the CIFAR10 dataset. Notably, {\pname} outperforms all other baseline methods, with SRAAL ranking as the second-best performing AL baseline. Initially, for smaller sample sizes, VAAL achieves comparable accuracy to other baseline methods but experiences a decline in performance in subsequent iterations, ultimately falling behind all other methods in comparison. MC-Dropout initially lags behind all other methods, but by the third iteration, it surpasses VAAL and achieves accuracy levels similar to Coreset. In summary, {\pname} exhibits a substantial $8.13\%$ higher mean accuracy compared to the best-performing baseline.

\textbf{Performance on CIFAR100.} We next present the results on the more challenging CIFAR100 dataset. In Figure~\ref{fig:balanced_datasets}(b), we present the empirical performance of active learning (AL) methods, including {\pname}, on CIFAR100. It's apparent that {\pname} emerges as the top performer, outperforming all other baselines by a significant margin. The recent approach, SRAAL, takes second place, although Coreset and LL4AL exhibit highly competitive results compared to SRAAL. In the final iteration, both MC-Dropout and VAAL achieve similar accuracy but lag behind all other compared baselines. Overall, {\pname} demonstrates a $1.79\%$ absolute improvement over the second-best-performing method, SRAAL.

\begin{figure*}[t]

  \centering
  \includegraphics[width=\textwidth, height=6.5cm]{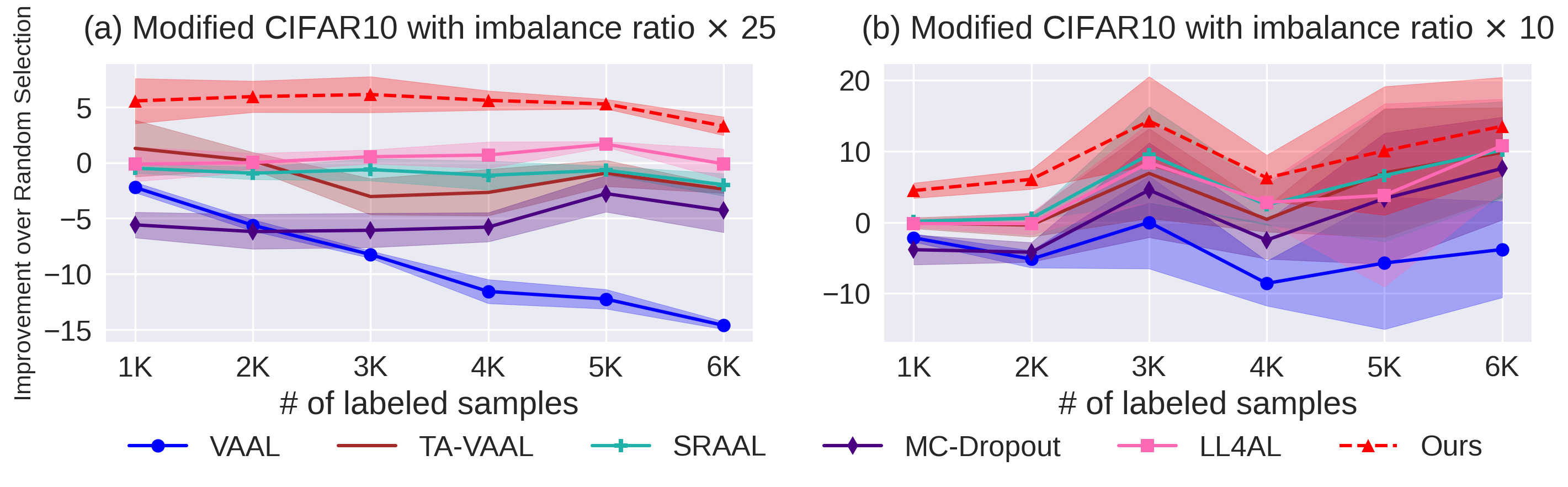}  

  \caption{Mean accuracy improvements with standard deviations (shaded) of AL methods w.r.t random sampling over the number of labeled samples on modified imbalanced CIFAR10 dataset.}
  \label{fig:imbalanced_cifar10}
\end{figure*}

\begin{wrapfigure}{R}{0.55\textwidth}
  \centering
  
  
  \includegraphics[width=8.6cm, height=7.4cm]{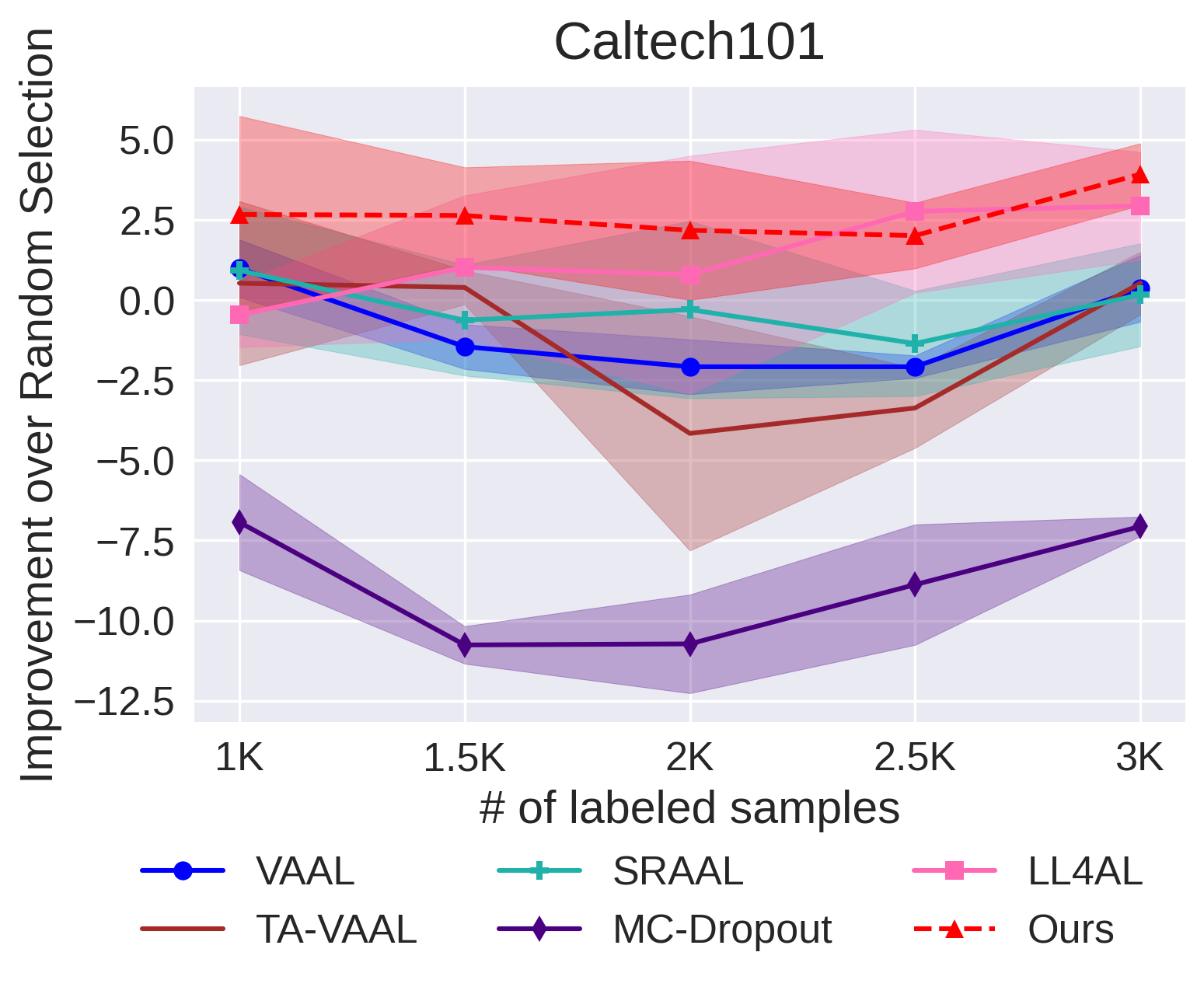}
  

  
  \caption{Mean accuracy improvements with standard deviations (shaded) of AL methods w.r.t random sampling over the number of labeled samples on Caltech101 dataset.}
  \label{fig:imbalanced_caltech101}
  
\end{wrapfigure}

\textbf{Performance on TinyImageNet-200.} TinyImageNet-200, a subset of ImageNet-1K~\citep{deng2009imagenet}, presents a more challenging task than CIFAR10/100. In Figure~\ref{fig:balanced_datasets}(c), we observe the empirical performance of the compared baselines. Notably, {\pname} outperforms all previous baselines, establishing its efficacy on this challenging dataset. In this setting too, SRAAL is the best performer in the baseline methods. Remarkably, VAAL attains better accuracy on TinyImageNet-200 compared to CIFAR10/100. VAAL even initially ranks as the second-best-performing AL method. Random sampling initially outperforms LL4AL and TA-VAAL but eventually converges to a similar level of performance as these methods. MC-Dropout consistently performs the worst among all baselines. Overall, {\pname} achieves a performance gain of $2.6\%$ with respect to SRAAL, showing the efficacy of the proposed model over the highly challenging dataset.

\textbf{Performance on ImageNet-100.} ImageNet-100 is a subset of ImageNet-1K~\citep{deng2009imagenet}, wherein the subset's classes are randomly selected. {Detailed information about the 100 classes is available in the appendix.} Figure~\ref{fig:balanced_datasets}(d) demonstrates the empirical performances of various active learning (AL) approaches, including {\pname}. Notably, {\pname} consistently outperforms all other baselines throughout the AL iterations, achieving a $3.6\%$ performance gain compared to the second-best AL method, VAAL. Remarkably, SRAAL underperforms compared to VAAL but outperforms other baselines. In the initial AL iterations, the other baselines (excluding MC-Dropout) achieve similar accuracy, but by the third iteration, TA-VAAL surpasses them, ultimately performing similarly to SRAAL. MC-Dropout, in contrast to other baselines, including {\pname}, exhibits the worst performance throughout the AL iterations on ImageNet-100. In summary, {\pname} emerges as the top-performing AL method on ImageNet-100, a subset of ImageNet-1K, with an overall $3.6\%$ higher mean accuracy.

\subsection{Active Learning on Imbalanced Datasets}\label{subsec:imbalanced_dataset}

\textbf{Datasets:} To evaluate the robustness of {\pname}, we test its performance on imbalanced datasets where class sample sizes significantly vary. We use two imbalanced datasets: $(i)$ a Modified Imbalanced CIFAR10 and $(ii)$ Caltech101~\citep{fei2006one}. In Modified Imbalanced CIFAR10, we randomly decrease sample sizes in the first five classes to introduce imbalance, with two ratios: $25$ and $10$. Caltech101 is inherently imbalanced, containing $8677$ images distributed across $101$ classes, with varying $(32-800)$ samples per class. For Caltech101, we randomly select $1000$ images with a $500$ budget for subsequent iterations. For Modified CIFAR10, we follow the same approach as CIFAR10, as described in Section~\ref{subsec:balanced_dataset}.

\textbf{Compared methods.} We evaluate the effectiveness of {\pname} in comparison to VAAL~\citep{sinha2019variational}, MC-Dropout~\citep{gal2017deep}, LL4AL~\citep{yoo2019learning}, SRAAL~\citep{zhang2020state}, TA-VAAL~\citep{kim2021task}, and Random Selection. The model initialization and training undergoes the same procedures as described for balanced datasets in Section~\ref{subsec:balanced_dataset}.

\textbf{Implementation details.} For the Modified imbalanced CIFAR10, we follow a similar implementation as described for CIFAR10 in Section~\ref{subsec:balanced_dataset}. We use a classical ResNet-18 architecture for Caltech101. We utilize a Modified Wasserstein autoencoder~\citep{tolstikhin2017wasserstein} as the VAE and employ a $5$-layer MLP as the discriminator. In both cases, we use the Adam optimizer~\citep{kingma2014adam} (lr: $5\times10^{-4}$). {For more details, please refer to the appendix.}

\begin{wrapfigure}{R}{0.55\textwidth}
  \centering
  
  
  \includegraphics[width=8.2cm, height=6.6cm]{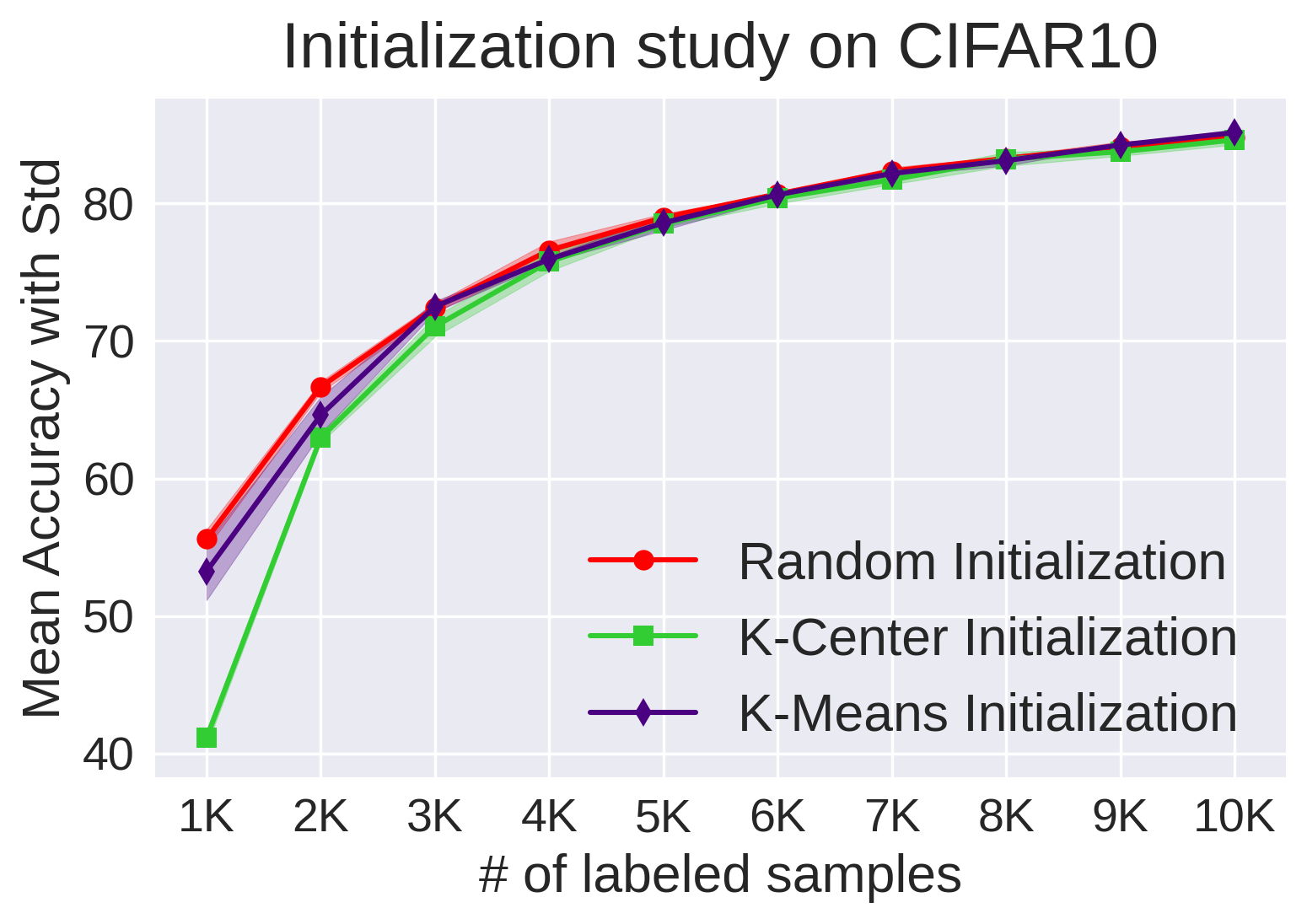}
  

  
  \caption{Comparison of initialization methods on the performance of {\pname} using (balanced) CIFAR10 dataset.}
  \label{fig:initialization_study}
  
\end{wrapfigure}

\textbf{Performance on modified CIFAR10.} Figure~\ref{fig:imbalanced_cifar10} illustrates the mean accuracy improvement with standard deviations relative to random sampling across various numbers of labeled samples on the modified CIFAR10 dataset. The dataset has two imbalance ratios, $10$ (more imbalanced) and $25$ (less imbalanced). {\pname} consistently outperforms all the other methods on both versions of modified imblanced CIFAR10. For an imbalance ratio of $25$, LL4AL follows as the second-best-performing AL approach. Meanwhile, in the case of an imbalance ratio of $10$, SRAAL initially achieves the second-best accuracy; however, by the end, LL4AL surpasses SRAAL to become the second-best-performing AL approach. For absolute mean accuracy with standard deviations, refer to the {appendix}.

\textbf{Performance on Caltech101.} In Figure~\ref{fig:imbalanced_caltech101}, we present the results for the various baselines along with the proposed model {\pname} on the highly imbalanced Caltech101 dataset. We show the metric mean accuracy improvement, along with standard deviations, in comparison to random sampling. We can observe that {\pname} consistently outperforms all other baseline methods by a significant margin. Initially, VAAL, SRAAL, and TA-VAAL achieve the second-highest accuracy; however, by the second iteration, their performance begins to decline. In contrast, LL4AL exhibits substantial improvement and ultimately emerges as the second-best performing AL method. MC-Dropout consistently performs the poorest among the compared baselines. For absolute mean accuracy values with standard deviations, please refer to the {appendix}.

\subsection{Comparison of initialization methods in Active Learning}\label{subsec:initialization_study}

In this section, we explore different initialization strategies for the initial labeled pool selection in our AL method, {\pname}, and assess their impact. While many AL methods use random initialization, a practice that we also follow;  we examine two other strategies: $(i)$ $K$-center-based sampling, and $(ii)$ $K$-means-based sampling. Figure~\ref{fig:initialization_study} depicts {\pname}'s performance under these strategies. Random sampling shows the best initial performance, followed by $K$-means-based sampling, with $K$-center-based sampling exhibiting the poorest initial performance. Ultimately, $K$-means-based sampling achieves the highest accuracy, suggesting its potential as an alternative to random sampling.

\section{Ablation}\label{sec:ablations}





\begin{figure*}[t]

  \centering
  \includegraphics[width=\textwidth, height=6.6cm]{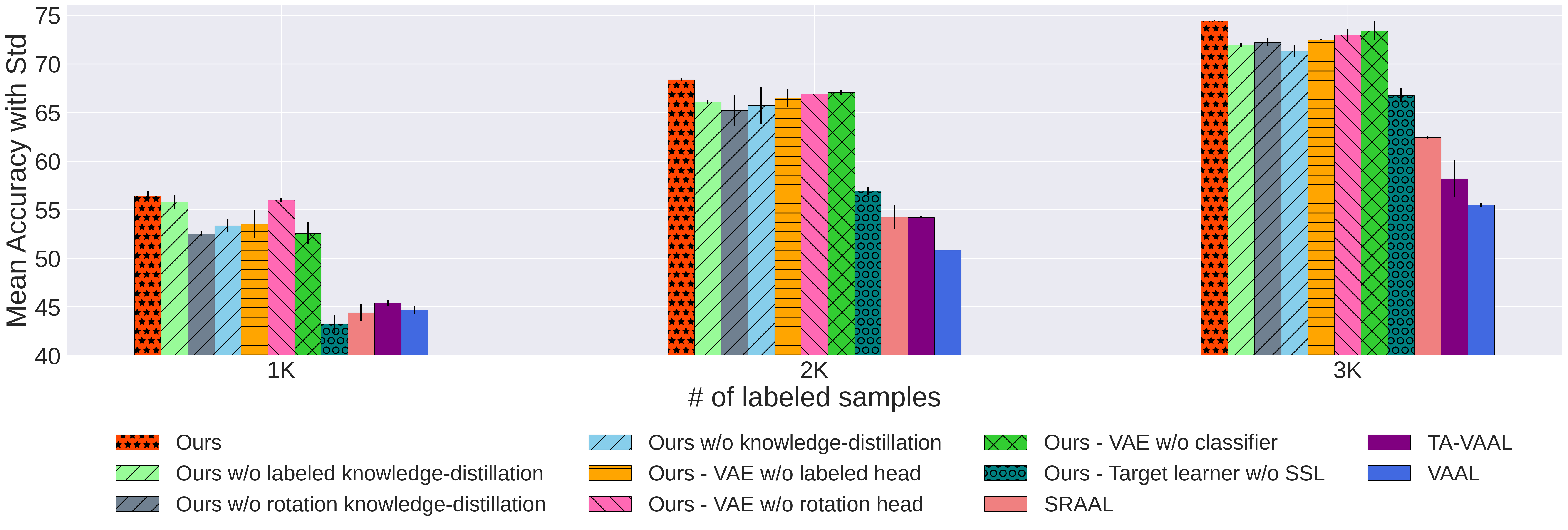}  

  \caption{Results of ablation studies by selectively removing core components of {\pname} on (balanced) CIFAR10.}
  \label{fig:ablaion_study}
\end{figure*}

We perform ablation experiments on {\pname}, assessing the impact of key components. {\pname} crucially integrates teacher-student learning and applies self-supervised learning to unlabeled data. Evaluation includes scenarios with or without knowledge distillation and self-supervised learning. Additionally, we systematically modify the classifier and target learner. Figure~\ref{fig:ablaion_study} depicts {\pname}'s performance with or without key components on CIFAR10. {\pname} consistently outperforms variations, demonstrating the synergistic effectiveness of our proposed components in selecting informative unlabeled samples. Even {\pname} without specific components outperforms existing baselines.

\section{Conclusion}\label{sec:conclusions}

This paper introduces \emph{A Self-Supervised Framework for Learning Robust Representations for Active Learning} ({\pname}), a novel task-aware active learning framework. {\pname} leverages both labeled and unlabeled data to detemine the most informative subset from the unlabeled pool. It establishes conditional relationships using labeled data through a proxy learner and enhances its performance through self-supervised learning on unlabeled data. Employing teacher-student learning between target and proxy learners ensures the proxy learner mimics the target-learner's behavior. Empirical evidence across diverse datasets, including CIFAR10, CIFAR100, TinyImageNet-200, ImageNet-100, and Caltech101, consistently demonstrates {\pname} outperforming state-of-the-art active learning methods. Furthermore, our investigation into sampling strategies identifies $K$-means-based sampling as a promising alternative to random sampling. Ablation studies validate the significance of the proposed components in {\pname}.

\bibliography{main}
\bibliographystyle{tmlr}

\clearpage

\appendix



  


  






\section{Additional Results: Active Learning on Imbalanced Datasets}

\begin{figure}[!htbp]

  \centering
  \includegraphics[width=\textwidth, height=7cm]{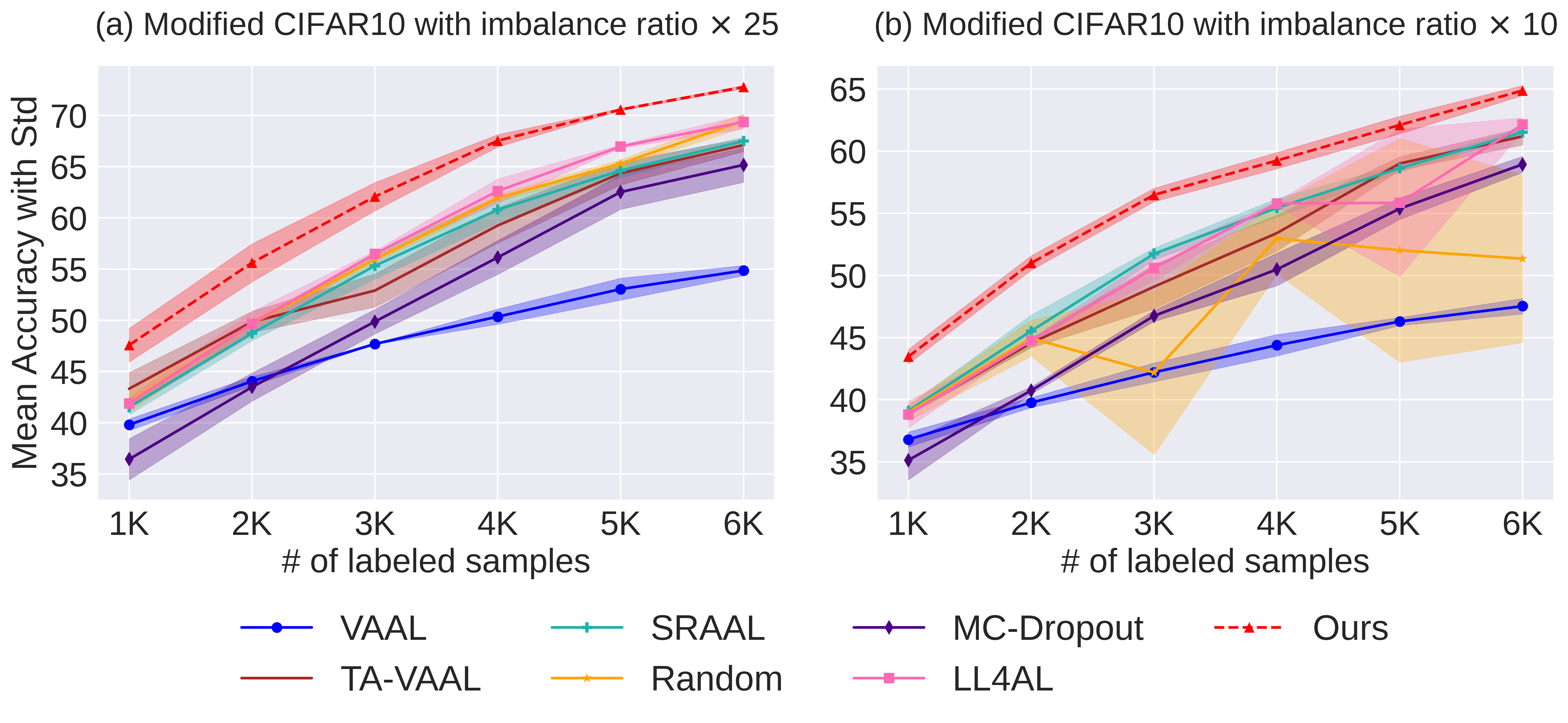}  

  \caption{Mean accuracy with standard deviations (shaded) of AL methods over the number of labeled samples on modified imbalanced CIFAR10 dataset.}
  \label{fig:imbalanced_cifar10_all}
\end{figure}

\begin{figure}[!htbp]

  \centering
  \includegraphics[width=0.75\textwidth, height=8.5cm]{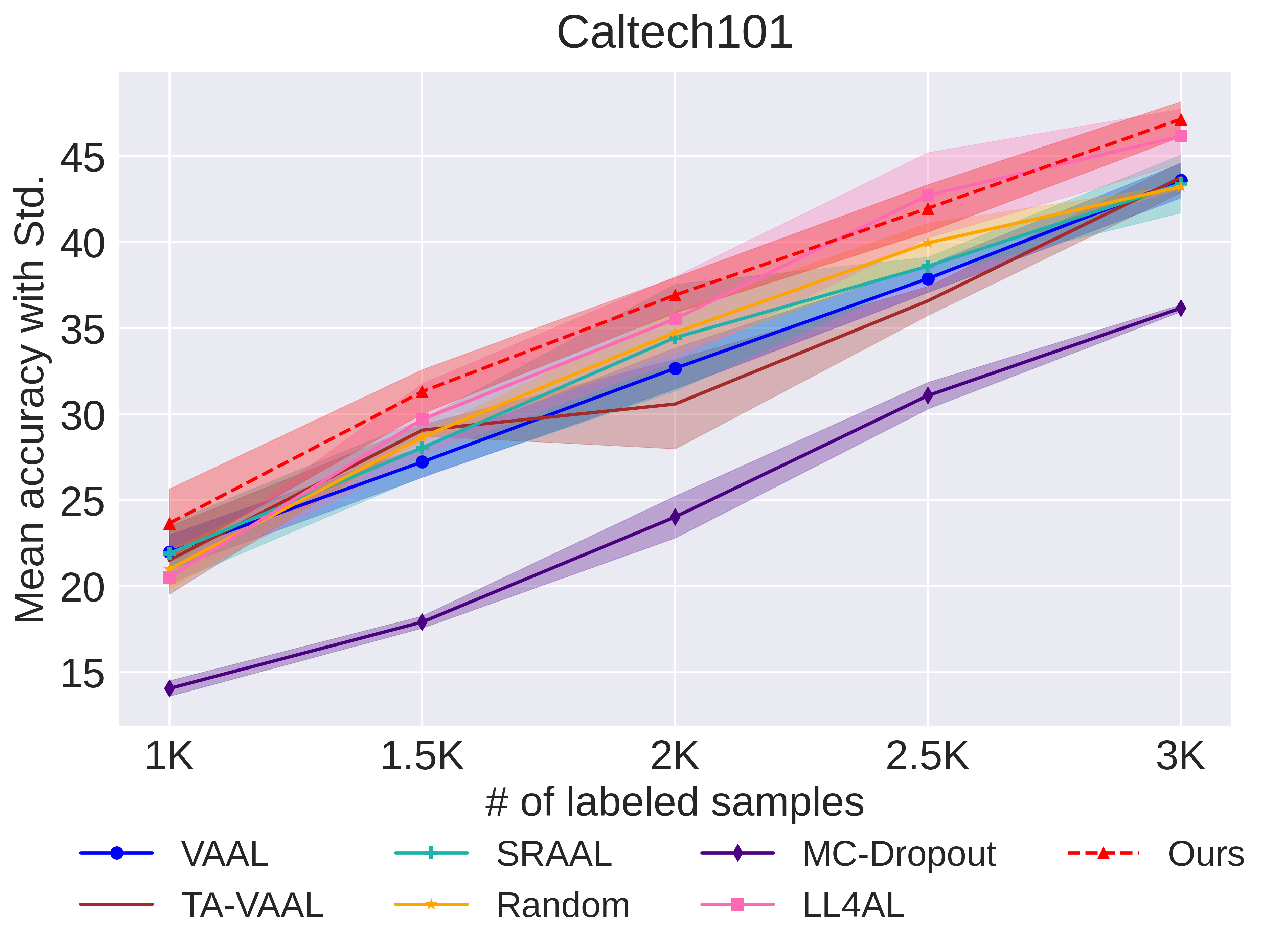}  

  \caption{Mean accuracy with standard deviations (shaded) of AL methods over the number of labeled samples on Caltech101 dataset.}
  \label{fig:imbalanced_caltech101_all}
\end{figure}

Figures~\ref{fig:imbalanced_cifar10} and \ref{fig:imbalanced_caltech101} in the main paper, show the performance improvement (mean accuracy with standard deviation) relative to Random sampling over the number of labeled samples on modified CIFAR10 and Caltech101 respectively. In this section, we provide the absolute accuracy curves over the number of labeled samples on modified CIFAR10 and Caltech101 in Figure~\ref{fig:imbalanced_cifar10_all} and \ref{fig:imbalanced_caltech101_all} respectively. 


\section{ImageNet-100}\label{imagenet100}

In this paper, we employed a subset of ImageNet-1K (ILSVRC-2012)~\cite{deng2009imagenet}, comprising randomly selected 100 classes. To facilitate a thorough investigation, we have provided the list of these 100 classes utilized for assessing the performance of the active learning baselines in our experiments, as detailed in Table\ref{Table_imagenet100_list}.

\section{Additional Ablation on CIFAR100}

In addition to the ablation study presented in Section~\ref{sec:ablations}, we also perform an ablation experiment on {\pname} utilizing the (balanced) CIFAR-100 dataset. This experiment assesses the impact of self-supervised learning, implemented by predicting the random transformations (Figure~\ref{fig:rotation_augmentations}) applied to the unlabeled samples, on the overall model performance.

\begin{figure}[!htbp]

  \centering
  \includegraphics[width=\textwidth, height=6.2cm]{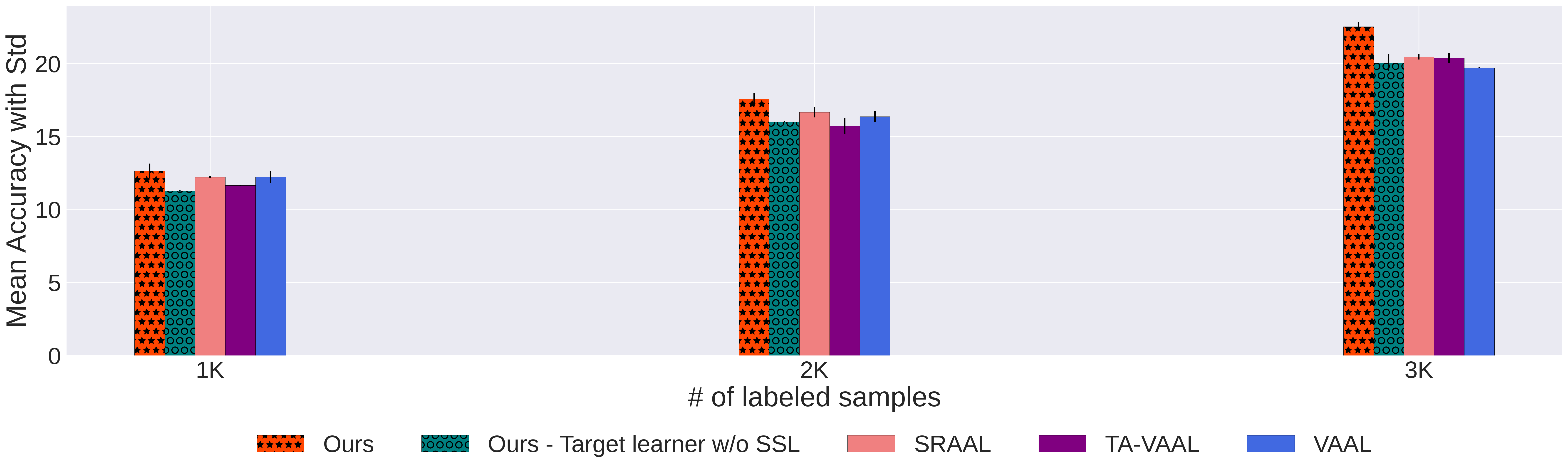}  

  \caption{Results of ablation study comparing the performance of {\pname} with or without using self-supervisied learning in target task-learner training on (balanced) CIFAR100.}
  \label{fig:ablaion_study_cifar100}
\end{figure}

Figure~\ref{fig:ablaion_study_cifar100} demonstrates the performance of {\pname} with or without applying self-supervised learning on the target task learner in comparison to existing active learning baselines. It can be seen that {\pname} with all the proposed components, including self-supervised loss on the target task learner, achieves the best performance compared to the existing active learning baselines. However, {\pname} without using self-supervision in target task learner training performs slightly worse compared to the existing active learning baselines. This highlights the importance of self-supervised learning in the overall model performance and the significance of exploiting the freely available abundant pool of unlabeled samples to achieve the best overall model performance. As future work, we will explore the effects of utilizing unlabeled samples in active learning model performance through various self-supervised and semi-supervised approaches.

\section{Details on hyperparameters}

Table~\ref{tab:hyperparameters} shows the hyperparameters that we use to train our proposed model ({\pname}). We set these hyperparameters following VAAL~\cite{sinha2019variational} and SRAAL~\cite{zhang2020state} settings.

\begin{table*}[t]
  \small

  \caption{The list of classes from ImageNet-100, which are randomly chosen from the original ImageNet-1K (ILSVRC-2012).} 
  \label{Table_imagenet100_list}

  \centering
  
  \addtolength{\tabcolsep}{10pt}
  \begin{tabular}{ c  c  c  c } 
    \toprule
  
    \multicolumn{4}{c} {\textbf{List Of ImageNet-100 Classes}} \\

    \midrule

    n01632777 & n01667114 & n01744401 & n01753488 \\
    
    n01768244 & n01770081 & n01798484 & n01829413 \\

    n01843065 & n01871265 & n01872401 & n01981276 \\
    
    n02006656 & n02012849 & n02025239 & n02085620 \\

    n02086079 & n02089867 & n02091831 & n02094258 \\

    n02096294 & n02100236 & n02100877 & n02102040 \\

    n02105251 & n02106550 & n02110627 & n02120079 \\

    n02130308 & n02168699 & n02169497 & n02177972 \\

    n02264363 & n02417914 & n02422699 & n02437616 \\

    n02483708 & n02488291 & n02489166 & n02494079 \\

    n02504013 & n02667093 & n02687172 & n02788148 \\

    n02791124 & n02794156 & n02814860 & n02859443 \\

    n02895154 & n02910353 & n03000247 & n03208938 \\

    n03223299 & n03271574 & n03291819 & n03347037 \\

    n03445777 & n03529860 & n03530642 & n03602883 \\

    n03627232 & n03649909 & n03666591 & n03761084 \\

    n03770439 & n03773504 & n03788195 & n03825788 \\

    n03866082 & n03877845 & n03908618 & n03916031 \\

    n03929855 & n03954731 & n04009552 & n04019541 \\

    n04141327 & n04147183 & n04235860 & n04285008 \\

    n04286575 & n04328186 & n04347754 & n04355338 \\

    n04423845 & n04442312 & n04456115 & n04485082 \\

    n04486054 & n04505470 & n04525038 & n07248320 \\

    n07716906 & n07730033 & n07768694 & n07836838 \\

    n07860988 & n07871810 & n11939491 & n12267677 \\
     
   \bottomrule
  \end{tabular}
  
  
\end{table*}



  
  

    

    

    













  
    
      

\begin{table*}[t]


  \scriptsize
  \centering
  \caption{Hyperparameters used in {\pname}. $d$ denotes the dimension of the latent space of the VAE. $\alpha_{1}, \alpha_{2}$ and $\alpha_{3}$ are the learning rates of VAE, Discriminator and Target task-learner respectively. $\lambda_{1}, \lambda_{2}, \lambda_{3}$ and $\lambda_{4}$ are the regularization hyperparameters that determine the effect of various components in learning an effective latent space representation using VAE, used in Eq. (7). $\beta$ is the Lagrangian parameter used in Eq. (1). `initial' denotes the size of inital labeled pool. `budget' denotes the number of unlabeled samples selected in the subsequent iterations.}
  \label{tab:hyperparameters}
  
  \resizebox{\textwidth}{!}{
  
    \small

    
    \begin{tabular}{ c | c | c | c | c | c | c | c | c | c | c | c | c | c } 

      \toprule
    
       & \multicolumn{13}{c}{Hyperparameters} \\

      \cmidrule{2-14} 
    
      \textbf{Dataset} & $d$ & $\alpha_{1}$ & $\alpha_{2}$ & $\alpha_{3}$ & $\lambda_{1}$ & $\lambda_{2}$ & $\lambda_{3}$ & $\lambda_{4}$ & $\beta$ & epochs & batch-size & initial / budget & image-size \\

      \midrule

      \shortstack{CIFAR10 \\ \cite{krizhevsky2009learning}} & 32 & $5 \times 10^{-4}$ & $5 \times 10^{-4}$ & $1 \times 10^{-2}$ & 1 & $5 \times 10^{-1}$ & $5 \times 10^{-1}$ & 1 & $1$ & $100$ & $128$ & $1000 / 1000$ & $32 \times 32$ \\

      \midrule

      \shortstack{CIFAR100 \\ \cite{krizhevsky2009learning}} & 32 & $5 \times 10^{-4}$ & $5 \times 10^{-4}$ & $1 \times 10^{-2}$ & 1 & $5 \times 10^{-1}$ & $5 \times 10^{-1}$ & 1 & $1$ & $100$ & $128$ & $1000 / 1000$ & $32 \times 32$ \\

      \midrule

      \shortstack{TinyImageNet-200 \\ \cite{le2015tiny}} & 32 & $5 \times 10^{-4}$ & $5 \times 10^{-4}$ & $1 \times 10^{-2}$ & 1 & $5 \times 10^{-1}$ & $5 \times 10^{-1}$ & 1 & $1$ & $100$ & $128$ & $2000 / 2000$ & $64 \times 64$ \\

      \midrule

      \shortstack{ImageNet-100 \\ \cite{deng2009imagenet}} & 128 & $5 \times 10^{-4}$ & $5 \times 10^{-4}$ & $1 \times 10^{-2}$ & 1 & $5 \times 10^{-1}$ & $5 \times 10^{-1}$ & 1 & $1$ & $100$ & $32$ & $2000 / 2000$ & $224 \times 224$ \\

      \midrule

      \shortstack{Modified \\ Imbalanced \\  CIFAR10} & 32 & $5 \times 10^{-4}$ & $5 \times 10^{-4}$ & $1 \times 10^{-2}$ & 1 & $5 \times 10^{-1}$ & $5 \times 10^{-1}$ & 1 & $1$ & $100$ & $128$ & $1000 / 1000$ & $32 \times 32$ \\

      \midrule

      \shortstack{Caltech101 \\ \cite{fei2006one}} & 128 & $5 \times 10^{-4}$ & $5 \times 10^{-4}$ & $1 \times 10^{-2}$ & 1 & $5 \times 10^{-1}$ & $5 \times 10^{-1}$ & 1 & $1$ & $100$ & $32$ & $1000 / 500$ & $224 \times 224$ \\

      \bottomrule
    \end{tabular}
  
  }
    
      
\end{table*}

\end{document}